**A Taxonomy of Generative AI in HEOR: Concepts, Emerging Applications, and Advanced Tools – An ISPOR Working Group Report**

**Rachael L. Fleurence, PhD[1], Xiaoyan Wang, PhD[2,3], Jiang Bian, PhD[4,5,6], Mitchell K. Higashi, PhD[7], Turgay Ayer, PhD[8,9], Hua Xu, PhD[10], Dalia Dawoud, PhD[11,12], Jagpreet Chhatwal, PhD[13,14]**

1. National Institute of Biomedical Imaging and Bioengineering, National Institutes of Health, Bethesda, MD, United States
2. Tulane University School of Public Health and Tropical Medicine, New Orleans, LA
3. Intelligent Medical Objects, Rosemont, IL, United States
4. Health Outcomes and Biomedical Informatics, College of Medicine, University of Florida, FL, United States
5. Biomedical Informatics, Clinical and Translational Science Institute, University of Florida, FL, United States
6. Office of Data Science and Research Implementation, University of Florida Health, Gainesville, FL, United States
7. ISPOR, The Professional Society for Health Economics and Outcomes Research, Lawrenceville, NJ, United States
8. Center for Health & Humanitarian Systems, Georgia Institute of Technology, Atlanta, GA,
9. Value Analytics Labs, Boston, MA, United States
10. Institute Department of Biomedical Informatics and Data Science, School of Medicine, Yale University, New Haven, CT, United States
11. National Institute for Health and Care Excellence, London, United Kingdom
12. Cairo University, Faculty of Pharmacy, Cairo, Egypt
13. Institute for Technology Assessment, Massachusetts General Hospital, Harvard Medical School, Boston, MA, United States
14. Center for Health Decision Science, Harvard University, Boston, MA, United States

**Funding:** Dr Dalia Dawoud reports partial funding from the European Union's Horizon 2020 research and innovation programme under Grant Agreement No. 82516 (Next Generation Health Technology Assessment (HTx) project. No other funding was received.

**Acknowledgements**: This manuscript was developed as part of the International Society for Pharmacoeconomics and Outcomes Research (ISPOR) Working Group on Generative AI. The authors wish to thank the ISPOR Science office for their support, Sahar Alam for her excellent program management throughout the project and Soo Tan for her assistance with the figure. The views expressed are those of the authors and do not necessarily reflect the official policy or position of their employers those of their employers or funding organizations.

**Highlights**

**What methods or evidence gap does your paper address?**
This article is an introduction to how generative AI can be effectively applied in HEOR. It introduces a taxonomy of AI terms, applications and tools and explores methods to enhance their accuracy and reliability. Key challenges, such as scientific validity and reliability, bias and fairness and technical and operational deployment, must be addressed before AI can be fully operational in HEOR.

**What are the key findings from your research?**
Generative AI shows promise in automating HEOR tasks, such as for systematic reviews, real-world evidence generation, economic modeling and dossier development. However, it is not yet reliable enough for autonomous use. Pairing AI with techniques like prompt engineering and retrieval-augmented generation can enhance the accuracy and dependability of results.

**What are the implications of your findings for healthcare decision-making or the practice of HEOR?**
Generative AI can improve HEOR processes but augment rather than replace human expertise in the near term. HEOR professionals and healthcare decision-makers should adopt generative AI tools, with strong checks and balances as its capabilities improve.

**Abstract**

**Objective:** This article presents a taxonomy of generative artificial intelligence (AI) for health economics and outcomes research (HEOR), explores emerging applications, outlines methods to improve the accuracy and reliability of AI-generated outputs and describes current limitations.

**Methods**: Foundational generative AI concepts are defined, and current HEOR applications are highlighted, including for systematic literature reviews, health economic modeling, real-world evidence generation, and dossier development. Techniques such as prompt engineering (e.g., zero-shot, few-shot, chain-of-thought, persona pattern prompting), retrieval-augmented generation, model fine-tuning, and domain-specific models, and use of agents are introduced to enhance AI performance. Limitations associated with the use of generative AI foundation models are described.

**Results:** Generative AI demonstrates significant potential in HEOR, offering enhanced efficiency, productivity, and innovative solutions to complex challenges. While foundation models show promise in automating complex tasks, challenges persist in scientific accuracy and reproducibility, bias and fairness and operational deployment. Strategies to address these issues and improve AI accuracy are discussed.

**Conclusion**: Generative AI has the potential to transform HEOR by improving efficiency and accuracy across diverse applications. However, realizing this potential requires building HEOR expertise and addressing the limitations of current AI technologies. Ongoing research and innovation will be key to shaping AI's future role in our field.

# A Taxonomy of Generative AI in HEOR: Concepts, Emerging Applications, and Advanced Tools – An ISPOR Working Group Report

## Introduction

Generative Artificial Intelligence (AI) is impacting multiple areas in science and medicine, including in health economics and outcomes research (HEOR)[1-3]. The field of AI has been exploring ways to use machine intelligence to augment human endeavors since the 1950s[4]. By the 1990s, machine learning (ML) techniques were advancing pattern recognition and decision-making processes[3]. By the 2000s, researchers had developed deep learning models based on neural networks, enabling a wide range of complex applications from image recognition to natural language processing (NLP)[3,5]. In 2021, a breakthrough in structural biology occurred when AlphaFold, a neural network-based deep learning program created by DeepMind, accurately predicted protein folding, significantly accelerating the process of drug discovery[6,7]. The scientists leading this effort were awarded a Nobel Prize in Chemistry in October 2024[8].

In the past few years, foundation models (FMs)—large-scale AI systems trained on extensive, unlabeled datasets through self-supervised learning—have emerged as transformative tools[9]. These models represent a significant shift in healthcare AI, transitioning from task-specific, single-purpose models to more versatile and adaptable generalist AI systems for medical applications[10,11]. A major paradigm shift occurred in November 2022 with the launch of OpenAI's ChatGPT, a type of generative AI that produces text, images, or other content based on input prompts[12,13]. Large Language Models (LLMs), a key technology behind FMs, can recognize, summarize, and produce coherent and contextually relevant texts[2]. In recent years, several major FMs have emerged, including OpenAI's GPT models, Google's Gemini, Anthropic's Claude, and Meta's Llama[14].

FMs have the potential to drive innovation in HTA domains like systematic literature reviews (SLRs), evidence synthesis, health economic modeling, real world evidence (RWE) generation, and value dossier development[3]. These models can streamline research processes and significantly enhance productivity. With these early applications in HEOR emerging, Health Technology Assessment (HTA) agencies are in the process of developing guidelines for the use of generative AI in submissions[3,15]. For instance, the National Institute for Health and Care Excellence (NICE) in the United Kingdom has issued a statement of intent[16] to describe their approach to AI in general as well as a position statement that covers the principles that should be adhered to when generative AI is used in submissions[15].

This manuscript is intended to serve as a resource for HEOR professionals exploring the rapidly evolving field of generative AI and its impact on HEOR workflows. As adoption grows, understanding key terms, techniques, and applications is crucial. In particular, the article introduces fundamental concepts like AI, ML, generative AI, and FMs (defined in **Box 1**, illustrated in **Figure 1**), highlights emerging HEOR applications, and reviews techniques to optimize generative AI use. It also examines current limitations, offering a balanced view of the opportunities and risks associated with the use of FMs in the field of HEOR.

This article is not a comprehensive review of all HEOR applications or tools but intends to serve as a starting point for understanding these technologies. It offers foundational knowledge and practical insights for assessing their relevance in HEOR. Designed for a broad audience, it provides accessible explanations for beginners and advanced insights into techniques and limitations for more experienced readers exploring AI integration into HEOR workflows.

**Figure 1: Relationship between AI, Gen AI, Foundation Models and LLMs**

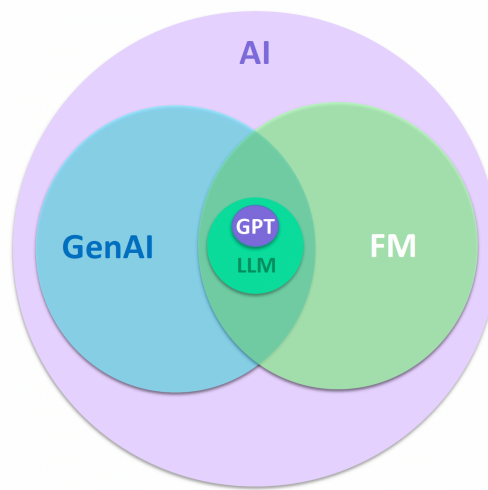

AI=Artificial Intelligence; FM= foundation model; Gen AI = generative artificial intelligence; GPT= Generative Pre-Trained Transformer; LLM= large language model.

**Examples of applications of generative AI in HEOR use-cases**

Applications using generative AI in HEOR are rapidly evolving, with use cases in SLRs, evidence synthesis, economic modeling, RWE generation, and dossier development[3]. **Table 1** describes the diverse applications of generative AI in HEOR showcasing its potential to streamline and enhance key processes in the field. The following section highlights key applications and associated challenges, drawing on the authors' experiences and a targeted literature review. While not exhaustive, it offers illustrative examples to guide HEOR professionals and showcase the diverse potential of generative AI.

*Applications of Generative AI to SLRs and Network Meta-Analyses (NMAs)*

An early exploration of generative AI and FMs has concentrated on SLRs, a critical process for evidence synthesis in health research[3]. SLRs are time-consuming and labor-intensive, requiring detailed and careful screening of abstracts and full text articles, bias assessment, precise and sometimes extensive data extraction, and may include quantitative meta-analyses involving a vast number of studies[17,18]. FMs can streamline this process in a range of SLR tasks. Specifically, FMs can assist in developing the literature search strategies and screening abstracts and full text articles for inclusion and exclusion using predefined criteria[19-22]. For example, a recent study demonstrated high accuracy using a GPT-4-based reviewer in PRISMA-based medical systematic literature reviews[23]. FMs can also provide reasoning for excluding certain abstracts and full text articles[24]. They can assist with bias assessment by applying a list of questions to full text articles[25,26]. These models can extract structured data from unstructured text from research papers[27]. They can be trained to identify key data points, such as population characteristics, interventions, comparators, and outcomes (PICOs), improving the speed of data extraction[28-30]. FMs can also generate code for running meta-analyses in R and Python, and other programming languages[30]. Additionally, they can generate concise summaries of the included studies, helping researchers synthesize findings more efficiently[30]. NMAs are widely used in HEOR to synthesize evidence from multiple studies and compare the relative effectiveness of interventions[31]. Recent advancements demonstrate the potential of FMs to streamline quantitative evidence synthesis by automating tasks such as extracting relevant study parameters, standardizing data inputs, and generating interpretable results [30,32]. For example, Reason et al. showed that GPT-4 was able to replicate data extraction in four network meta-analyses with an accuracy exceeding 99% [30]. More generally, the application of FMs in automating statistical analyses represents a transformative opportunity for HEOR [33,34].

FMs show promise in automating SLR tasks but still face limitations. There have been reports of hallucinations, where FMs generate plausible but incorrect mesh terms of fabricated citations[35]. Additionally abstract disposition and data extraction is not consistently accurate and require manual validation[25]. While some studies demonstrate FM performance comparable to human tasks, this is not consistently reliable, emphasizing the need for human oversight [35]. Until robust methods and standards for evaluating generative AI outputs are established, manual verification remains essential [3,15].

*Applications of Generative AI to Health Economic Modeling*

FMs can be used for a range of different applications in health economic modeling and have the potential to transform how models are conceptualized, developed, and utilized[3]. FMs can efficiently summarize existing economic models, providing a rapid synthesis of their methodologies and outcomes[36]. This capability is helpful for model parameterization and data extraction, where FMs can expedite the identification and integration of relevant data points[37]. FMs can assist with the creation of new or de novo health economic models by leveraging extensive amounts of existing literature and data[38]. For example, one study demonstrated that a FM could replicate a three-state partition model for non-small cell lung cancer and renal cell carcinoma[39]. Another proof-of-concept study showed that an FM could fully replicate a published 'simple' health economic model evaluating the cost-effectiveness of combination therapy for HIV infection, including extraction of model structure, parameters identifications, code development, and results evaluation[40].

FMs have the potential to assist with various higher complexity tasks in model development, though we are not aware of any published studies in this area at the time of writing. For instance, FMs could aid in validating existing models by cross-verifying assumptions and outputs against new data or parallel models. They might also support adapting models to different geographic or demographic contexts, improving their accuracy and applicability. Additionally, FMs could streamline platform transitions, such as converting models from Excel to R Shiny, by automating and error-checking the process. One of the more resource-intensive applications is conducting structural uncertainty analysis[41]. Here FMs could automate parts of the workflow, significantly reducing labor and time requirements. These current and potential applications highlight the ability of FMs to accelerate health economic modeling. However, human expertise and oversight remain essential, as standards for ensuring the accuracy and reliability of AI-generated models are still evolving[3,15,42,43].

*Applications of Generative AI to RWE Generation*

Generative AI and FMs have the potential to assist in generating RWE by improving efficiency, accuracy, and scale of real-world data that might be available for research. Only a small portion of electronic health records (EHR) data is structured and in a format that lends itself to statistical analysis with minimal processing[44]. Much detailed patient information is embedded in clinical documents and reports, which are in unstructured text. This unstructured data requires additional processing before it can be included in analytical datasets. For decades, advances in AI, particularly in natural language processing (NLP), have notably accelerated data extraction, information retrieval and summarization for RWE generation[45]. Emerging applications include deploying generative AI tools to extract information from unstructured electronic EHRs with the potential to accelerate their integration of this data into analyzable datasets, reduce manual efforts and minimize human errors[46,47]. For example, FMs were shown to be successful in extracting biomarker testing details from EHR documents[48]. However some studies have questioned the accuracy of FMs in their ability to map descriptive text to medical codes, with one study finding that accuracy remains below 50%[49]. However, this study did not use any fine-tuning which may have improved the results. Approaches to improve the accuracy of mapping unstructured to structured text include the use of specialized models, such as GatorTron, NYUTron and Me Llama which are trained using large clinical texts[50-52,53], and improving the prompts provided to GPT-3.5 and GPT-4[46]. Additionally, by integrating multimodal data sources (e.g. adding imaging and genomics' information as well as structured clinical data and unstructured texts), FMs might assist in providing more comprehensive insights to a wider set of healthcare problems[54,55]. For example, FMs could accurately forecast COVID-19 cases and hospitalizations using real-time, complex, non-numerical information—such as textual policies and genomic surveillance data—previously unattainable in traditional forecasting models[56]. Generative AI tools show significant promise in enhancing RWE generation by streamlining the processing of unstructured data and integrating diverse data sources. While challenges like accuracy and reliability persist, advancements in fine-tuning, specialized models, and multimodal approaches suggest a path toward more comprehensive and accurate datasets. This work is needed to improve the comprehensiveness of value assessments for healthcare interventions.

*Applications of Generative AI to Dossier Development and Reporting*

Generative AI and FMs can be used to enhance the efficiency of dossier development for pharmaceutical product reporting and submissions to HTA agencies[15]. FMs excel at writing, can follow instructions for required styles, and can mimic the same style provided in other documents[57,58]. By automating the collation and presentation of evidence, generative AI might reduce the time and resources needed to produce comprehensive reports. Additionally, since FMs are language-agnostic, they can generate documents suitable for different countries, facilitating international submissions and communications. FMs can also tailor communication materials to specific stakeholders, creating customized messages and visual aids that effectively convey the results of HEOR studies[59]. This makes them useful for example in developing lay summaries of technical reports[15]. Users must consider several limitations for effective FM implementation in dossier development. A primary concern is accuracy, as FMs can generate plausible but incorrect information (e.g., hallucinations), particularly with complex or nuanced scientific data. These inaccuracies risk undermining the credibility of submissions to regulatory agencies or HTA bodies, emphasizing the need for rigorous review processes to mitigate such risks.

*Evaluating research output using generative AI*

Generative AI holds significant promise for HEOR, but most applications remain in early developmental stages. Of these, systematic literature reviews have progressed the furthest beyond proof-of-concept, while others require further validation to ensure reliability and practical utility. At this stage, it is difficult to separate limitations stemming from user expertise from those intrinsic to generative AI tools. Output quality often depends on user interaction, particularly in tasks like systematic literature reviews or economic modeling. For instance, better prompting can enhance results, making user expertise a critical factor [39]. Therefore, it might be more practical to assess output quality using independent metrics that evaluate overall performance rather than separating user influence from the model's capabilities. To support this, evaluation frameworks are under development and will likely be helpful for both authors and reviewers[42,43].

## Techniques to improve the use of generative AI in HEOR

This section examines several more advanced techniques to enhance generative AI performance in HEOR. Strategies like prompt engineering and retrieval-augmented generation (RAG) improve accuracy, factuality, and comprehensiveness. Model fine-tuning and domain-specific FMs ensure contextually relevant outputs for specialized tasks. These methods, applicable across use cases, help address key challenges in accuracy and reliability. Table 2 summarizes key approaches to enhance the quality of generative AI outputs in HEOR.

*Prompt Engineering*

Prompt engineering involves crafting specific inputs to optimize the quality of FM outputs [60]. It includes designing instructions, or prompts, that guide FMs to produce specific outputs through various strategies such as zero-shot, few-shot, chain-of-thought, and persona pattern prompting[61,62,63]. Zero-shot prompting enables FMs to respond to novel queries based solely on the question posed without providing any examples ("zero"), while few-shot prompting enhances accuracy by offering a few relevant examples ("few shot") [64]. Chain-of-thought prompting facilitates complex decision-making by guiding the model to display its reasoning process step-by-step [65]. Persona pattern prompting tailors outputs to align with the expertise and style expected of particular professional personas like health economists or policy makers [60]. **Box 2** provides some examples of different types of prompts. Several prompt engineering techniques have been deployed in the conduct of systematic literature reviews and economic modeling[22,29,30] [39]. However, prompt engineering is not without limitations. Zero-shot and few-shot prompting may miss important nuances in complex tasks, chain-of-thought can generate overly verbose outputs, and persona pattern prompting requires accurate replication of professional personas which can be challenging[60].

*Model Fine Tuning*

Fine-tuning is a specialized technique where a pre-trained FM undergoes additional training using targeted datasets to refine its capabilities for specific tasks[66]. Fine-tuning might also take the shape of instruction tuning, using high-quality instruction-response pairs (i.e. pairs of questions and answers, known to be correct) [67,68]. Fine-tuning can significantly improve the model's performance on niche or complex tasks by adjusting its parameters to better reflect the unique needs of the application[48]. Self-improving feedback loops and reinforcement learning from human feedback (RLHF) is a form of fine-tuning[67]. Self-improving feedback loops iteratively refine prompts based on the outputs received, at the request of the model, to enhance model performance over time. RLFH also has limitations which are described in detail in this article[69]. Fine-tuning is different from training a specialized model from scratch. An example in HEOR is Bio-SIEVE, an FM designed to automate systematic reviews, specifically title and abstract screening. Instruction-tuned on Llama and Guanaco models, Bio-SIEVE classifies studies for inclusion or exclusion based on predefined criteria and provides reasoning for exclusion, improving the efficiency of systematic reviews across medical domains [24]. While fine-tuning offers significant benefits, it is not without challenges. It can be resource-intensive, requiring substantial computational power and expert knowledge to perform the fine-tuning.

There is also a risk of overfitting to the specific data used for training, which might limit the model's generalizability to other tasks[70].

*Domain-specific FMs*

Domain-specific FMs are tailored for particular tasks or domains, like biomedical research or healthcare by utilizing domain-specific datasets to train the FM in more specific domains[71]. For example, in healthcare, these models can be trained on clinical notes from EHRs and other biomedical literature and texts, enhancing their performance on tasks like clinical named entity recognition, reasoning tasks, question and answering, and text summarization. Domain-specific FMs can be trained from scratch using domain-specific data. For example, NYUTron, a large language model trained on unstructured clinical notes to predict important clinical outcomes, such as hospital mortality and length of stay, showed improvement compared to traditional models[51]. GatorTron and GatorTronGPT are a generative clinical FM developed using GPT-3 architecture and trained on clinical text from clinical departments and patients at the University of Florida Health[50,52]. Both NYUTron and GatorTron have been shown to achieve superior performance in clinical NLP tasks, like named entity recognition, compared to their general domain counterparts. In addition, domain specific FMs can be built by continuously pre-training of open-domain FMs (e.g. Llama) using domain specific data. Examples of specialized models in this category include Me-LLaMA[53] and BioClinicalBERT, which are based on Llama and BERT respectively and are continuously trained on biomedical and clinical texts[72]. The enhanced performance of domain-specific FMs comes with limitations, including high training costs and technical demands, which may be prohibitive for smaller organizations. However, some models, like GatorTron, are freely available to researchers under a license agreement[50]. Researchers should carefully balance the benefits of improved performance against the additional burdens of cost, learning time, and required expertise.

*Retrieval-Augmented Generation (RAG)*

Retrieval-Augmented Generation (RAG) is a sophisticated method that combines the broad knowledge base of FMs with precise, domain-specific data retrieval [73]. Conceptually a RAG system retrieves more up-to-date information, or task specific information from external knowledge or data sources than the FM was pre-trained on[74]. For example, a generative AI solution can employ RAG to verify facts by accessing external databases or websites in real-time, ensuring responses not only draw from a vast internal dataset but are also cross verified with the latest external references[75]. This capability significantly enhances the accuracy and reliability of the information provided, especially in rapidly evolving fields where the FM might not have been trained on the most up to date data[74,76]. Unless carefully managed, a limitation of RAG could be its lack of ability in handling conflicting information retrieval which can detrimentally affect RAG's output quality and is an active area of research [73,76,77].

*Agent-Based Techniques in Generative AI for HEOR*

Agents in generative AI are intelligent systems designed to perform specific tasks autonomously by combining a LLM or other FMs with additional capabilities like memory, task execution, interaction with external tools or data sources, and even collaboration with other agents[78,79]. Unlike standalone LLMs, which often handle individual language tasks (e.g., text summarization), agents coordinate multiple components to execute complex workflows. For instance, virtual assistants like Alexa or Siri powered with generative AI can now act as agents by responding to commands and performing tasks such as answering questions using integrated tools and external data sources [80]. A pioneering and popular programming framework for building LLM-based applications is LangChain, an open-source tool that connects LLMs to external resources [81]. LangChain also enables the creation of task-specific agents capable of automating workflows such as retrieving data, synthesizing information, or running analyses. LangGraph, a recent addition to LangChain, enables stateful, multi-actor LLM applications, supporting complex, interactive AI systems with planning, reflection, reflexion, and multi-agent coordination. Several other emerging agentic AI frameworks are gaining popularity, such as Microsoft's AutoGen[82], an open-source tool designed for building AI agents and enabling cooperation among multiple agents. These tools allow agents to perform HEOR tasks more efficiently by interacting with external systems and retaining task context.

In HEOR, AI agents have significant potential to automate time-intensive tasks. For instance, in evidence synthesis, agents can retrieve and summarize research findings, streamlining systematic literature reviews. In economic modeling, they can extract and organize cost and utility data from real-world evidence, accelerating data integration. Agents can also aid dossier development by automating data collation and report formatting, reducing manual effort and enhancing consistency. For example, GPT Researcher is, an autonomous agent, conducts comprehensive web research based on text prompts, producing detailed, factual reports with citations[83]. However, challenges remain, including the technical expertise needed for development, ensuring transparency and traceability of outputs, and addressing risks like errors or hallucinations. While agents show promise for HEOR, they are still in early development and require further refinement to ensure reliability and scalability[78,79].

**Limitations of Generative AI as applied to HEOR**

While generative AI and FMs present promising opportunities in HEOR, several limitations must be acknowledged to provide a balanced understanding about the appropriate use of these technologies.

**Scientific Rigor in Generative AI: Accuracy and Reproducibility**

The validity and reliability of generative AI tools in scientific research depend on their accuracy and reproducibility. Accuracy ensures outputs are factual, error-free, and comprehensive[42], while reproducibility allows findings to be independently verified under consistent conditions, building confidence in AI-generated insights [84,85]. Both are crucial for maintaining the credibility and utility of these tools in HEOR. Instances of inaccuracies in FM outputs, such as hallucinations—factually incorrect or fabricated information—have been documented, including examples like non-existent citations generated during literature searches[19,35]. In the RWE space, one study reported less than 50% accuracy when mapping unstructured text to correct codes [49]. Similarly, Chhatwal et al. found significant variability in how FMs represented disease progression in a Markov model for hepatitis C[37]. To improve accuracy, strategies such as refining prompts and employing RAG have been effective in reducing hallucinations and increasing contextual accuracy [77,86]. Domain-specific FMs like GatorTron and MeLLAMA, trained on large clinical datasets, and fine-tuning with domain-specific data have also enhanced performance in specialized research tasks. [52,55-57,87]. Reproducibility, a cornerstone of scientific research, remains a challenge due to the black-box nature of FMs. AI models learn patterns from vast datasets, introducing variability in outputs based on input changes. Ensuring reproducibility will require open sharing of data, code, and results, as well as adopting standardized reporting and transparency practices [84,85]. However, the probabilistic nature of these models means some variability will persist. HEOR-focused evaluation frameworks and reporting standards offer promising paths to improving both accuracy and reproducibility, providing structured guidance for the use of generative AI in HEOR[42,43].

**Bias and Fairness in Generative AI**

Bias and fairness are important considerations in the development and application of generative AI tools, including in HEOR[3]. Bias refers to systematic differences introduced during model development or deployment that can perpetuate inequities, while fairness ensures equitable operation across diverse populations and contexts [88-91]. FMs can propagate or amplify biases arising from stages such as training data selection or deployment, potentially causing harm to individuals and communities [89,90,92]. An interesting example highlighting concerns associated with bias comes from a study by Giyocha et al., where an AI model accurately predicted a patient's self-reported race from medical imaging across various modalities—an ability beyond clinical experts [3,93]. However, the study could not determine the specific features the model used, raising ethical questions about fairness. For instance, if models implicitly use race as a factor, they risk biased treatment recommendations, potentially exacerbating health disparities. Several strategies have been proposed to manage the risk of bias in FM outputs. Published surveys explore methodologies to evaluate fairness, identify sources of bias, and implement corrective actions [88,89,91,94,95]. These approaches include fairness metrics like demographic parity and equalized odds, bias audits, adversarial debiasing, reweighting training data, and fine-tuning with domain-specific datasets [88,89,91]. Synthetic datasets have also shown promise in improving fairness[96]. Research on evaluating and mitigating bias in FMs applied to HEOR are an active area of research [3,95].

**Technical and Operational Considerations in Generative AI for HEOR**

The integration of FMs into HEOR is not without technical and operational challenges, including compliance with national regulations, ensuring security and privacy, balancing open-source and proprietary models, addressing deployment issues, managing costs, and integrating FMs into existing workflows.

First, generative AI tools must comply with national and local regulations, such as the EU AI Act, HIPAA in the U.S., and GDPR in the EU[97]. FMs can pose risks by memorizing and reproducing sensitive data, including Protected Health Information (PHI) [98,99]. Safeguards like encryption, anonymization, access controls, and compliance with standards such as the Federal Information Security Management Act (FISMA) and the HITRUST Framework are critical to mitigate these risks [100,101]. Federated analytics offers a promising solution by enabling analysis across multiple sources without centralizing data, preserving privacy while ensuring analytical rigor [102].

Second, the choice between open-source and proprietary FMs plays a crucial role in shaping their adoption and use within HEOR [9]. Open-source models like GPT-Neo, LLaMA, and DeepSeek offer transparency and customization but require substantial resources, expertise, and maintenance to keep pace with the field. Proprietary models, such as GPT-4, are easier to deploy and optimized for performance but come with high costs, vendor lock-in, privacy concerns, and limited transparency, complicating bias and fairness assessments [103].

Third, deployment strategies significantly impact FM adoption. Cloud-based solutions offer scalability and accessibility but pose risks for compliance with regulations like HIPAA and GDPR due to potential data breaches [104,105]. On-premises hosting provides better data privacy control but demands significant investments in infrastructure, personnel, and maintenance. Both approaches face cost challenges, including licensing fees, infrastructure expenses, and customization expertise. Organizations must assess their needs and resources to choose secure and cost-effective strategies.

Finally, integrating FMs into HEOR workflows involves both technical and operational challenges. Platforms and tools like LangChain or API-based systems offer the building blocks to automate tasks such as systematic reviews or economic modeling but require advanced expertise and are often hindered by the current absence of standardized APIs and interoperability protocols[81]. Simplifying deployment tools and creating user-friendly interfaces, or mature products tailored to specific HEOR tasks, will be important for wider adoption. Organizational barriers, including resistance to change, reliance on traditional methods, limited AI expertise, and high implementation costs, further hinder adoption. Providing training and upskilling opportunities for HEOR professionals will be essential to overcoming these challenges and enabling the effective integration of FMs into HEOR workflows[106].

**Conclusion**

This article introduces key concepts in generative AI for the HEOR community, providing a foundation for understanding and engaging with this transformative technology. Generative AI and FMs offer significant potential to enhance HEOR by streamlining workflows, automating complex tasks, and introducing innovative solutions. However, this promise comes with challenges, including ensuring scientific accuracy and reproducibility, addressing bias and fairness, and managing technical and operational complexities. Advanced techniques such as prompt engineering and retrieval-augmented generation may mitigate some of these limitations, but the field is still in its early stages of optimizing these tools. For HEOR professionals, the path forward requires embracing generative AI with a commitment to rigorous validation, interdisciplinary collaboration, and ongoing learning. Thoughtfully integrating these tools can support equitable and impactful healthcare decisions, ultimately improving patient outcomes.

**Glossary (Adapted from [3])**

- **Artificial intelligence (AI)**: a broad field of computer science that aims to create intelligent machines capable of performing tasks typically requiring human intelligence.

- **Deep Learning**: a subset of machine learning algorithms that uses multilayered neural networks, called deep neural networks. Those algorithms are the core behind the majority of advanced AI models.

- **Foundation Model**: a large scale pretrained models that serve a variety of purposes. These models are trained on broad data at scale and can adapt to a wide range of tasks and domains with further fine-tuning.

- **Generative AI**: AI systems capable of generating text, images, or other content based on input data, often creating new and original outputs.

- **Generative Pre-trained Transformer (GPT)**: a specific series of FMs created by OpenAI based on the Transformer architecture, which is particularly well-suited for generating human-like text.

- **Large Language Model:** a specific type of FM trained on massive text data that can recognize, summarize, translate, predict, and generate text and other content based on knowledge gained from massive datasets.

- **Machine learning (ML)**: a field of study within AI that focuses on developing algorithms that can learn from data without being explicitly programmed.

- **Multimodal AI**: an AI model that simultaneously integrates diverse data formats provided as training and prompt inputs, including images, text, bio-signals, -omics data and more.

- **Prompt:** the input given to an AI system, consisting of text or parameters that guide the AI to generate text, images, or other outputs in response.

- **Prompt engineering**: creating and adapting prompts (input) to instruct AI models to generate specific output.

- **Supervised Learning:** A machine learning approach where models are trained on labeled data, pairing inputs with known outputs. This enables the model to learn patterns for tasks such as predicting healthcare costs, diagnosing diseases, or classifying images.
- **Token:** A unit of text processed by an AI model, which can be a word, subword, or character. AI models convert input text into tokens to generate or interpret language. The number of tokens impacts model performance, cost, and the ability to process longer texts efficiently.
- **Unsupervised Learning:** A machine learning approach where models are trained on unlabeled data to uncover hidden patterns or structures, such as clustering patients with similar health outcomes.

**Box 1: Key Definitions**

*Artificial Intelligence (AI)*

Artificial intelligence refers to the ability of machines to perform tasks that typically require human intelligence, such as pattern recognition, language understanding, reasoning, and decision-making[4]. In HEOR, AI is increasingly used to automate complex and time-intensive tasks, including data extraction, statistical analysis, evidence synthesis, and predictive modeling, offering opportunities to enhance efficiency and accuracy in research.

*Machine Learning (ML)*

Machine learning, an important subset of AI, allows algorithms to learn from data to perform tasks without explicit programming [4,107]. ML enables computers to adapt and improve their ability to solve problems over time. ML began to take root in the 1990s with the development of groundbreaking techniques such as Support Vector Machines (SVM) and Random Forests[108,109]. ML has been used in HEOR to assist with various research tasks[110].

*Generative AI*

Generative AI represents a more advanced application of AI, capable of creating new content, synthesizing data, and providing innovative solutions to complex problems[3]. Generative AI models, particularly FMs, can process and produce natural language text, perform tasks that require reasoning, generate computer code, summarize research findings, draft reports, and much more[2,12,14]. These models are continuously improving and can be deployed to assist in various HEOR tasks, potentially providing greater accuracy and efficiency in research[3].

*Foundation Models (FMs)*

FMs are large-scale AI systems trained on massive datasets using self-supervised learning, enabling them to generalize across a wide range of tasks[2,9,12]. They are central to generative AI, offering versatility by performing well across multiple domains with minimal fine-tuning. Examples include GPT-4, Gemini, Claude, and LLaMA, used in applications ranging from text generation to image and video analysis[14]. Their adaptability makes them ideal for tasks like information extraction, language understanding, text summarization, and visual data interpretation.

*Large Language Models (LLMs)*

Large language models (LLMs) are a specialized type of FM designed for processing and generating human language[14]. Trained on extensive text corpora, LLMs excel at tasks like summarization, question answering, and content generation. Examples such as GPT-4, Claude, and LLaMA demonstrate competitive performance across diverse language tasks, often rivaling task-specific models with minimal fine-tuning[46]. While all LLMs are FMs, they focus exclusively on language-based applications.

**Box 2: Prompting Examples for Systematic Review of Treatments for Hepatitis C Virus**

**Zero-shot prompt:**

"Screen the following abstracts and classify them as relevant or irrelevant to the systematic review of treatments for Hepatitis C Virus (HCV), based on study type, population, and outcomes. Provide a binary response for each abstract: relevant or irrelevant."

**Few-shot prompt:**

"Here are a few examples of abstracts that have already been classified. Use these examples to guide your classification of new abstracts for the HCV treatment review.

  - Example 1: This randomized controlled trial evaluates the effectiveness of a new antiviral drug for HCV in patients with genotype 1. The study includes relevant clinical outcomes such as sustained virologic response (SVR) and adverse events. → Relevant

  - Example 2: This study examines the prevalence of HCV in a specific geographic region without discussing treatment outcomes. → Irrelevant

  - Now, based on these examples, classify the following abstracts as relevant or irrelevant to the systematic review."

**Chain-of-Thought prompt:**

 "Read the following abstract and explain your reasoning step-by-step before making a classification decision for the systematic review on HCV treatments. Focus on the type of study, the population studied, and whether the abstract includes treatment outcomes relevant to HCV care."

  - Example Abstract: This observational study analyzes the long-term efficacy of a novel direct-acting antiviral (DAA) treatment in HCV patients across multiple genotypes. Outcomes include Sustained Virologic Response (SVR) and patient-reported quality of life measures.

 The FM might provide an answer like this: "The study is observational, which can be included in a systematic review depending on the outcomes. The focus is on HCV treatment and includes important clinical outcomes like SVR and quality of life, which are directly relevant to the review question."

**Persona Pattern Prompting:**

"You are an experienced health economist tasked with identifying studies for a systematic review on HCV treatments. Your goal is to prioritize studies with robust methodologies that report clinical outcomes such as SVR and side effects. Classify the following abstract as relevant or irrelevant, considering your expertise in evaluating economic and clinical evidence."

**References:**

1. The Economist. AIs will make health care safer and better. *The Economist*. March 27 2024 2024;
2. Telenti A, Auli M, Hie BL, Maher C, Saria S, Ioannidis JPA. Large language models for science and medicine. *Eur J Clin Invest*. Feb 21 2024:e14183. doi:10.1111/eci.14183
3. Fleurence RL, Bian J, Wang X, et al. Generative AI for Health Technology Assessment: Opportunities, Challenges, and Policy Considerations - an ISPOR Working Group Report. *Value Health*. Nov 11 2024;doi:10.1016/j.jval.2024.10.3846
4. Howell MD, Corrado GS, DeSalvo KB. Three Epochs of Artificial Intelligence in Health Care. *JAMA*. Jan 16 2024;331(3):242-244. doi:10.1001/jama.2023.25057
5. Hinton GE, Osindero S, Teh YW. A fast learning algorithm for deep belief nets. *Neural Comput*. Jul 2006;18(7):1527-54. doi:10.1162/neco.2006.18.7.1527
6. Jumper J, Evans R, Pritzel A, et al. Highly accurate protein structure prediction with AlphaFold. *Nature*. Aug 2021;596(7873):583-589. doi:10.1038/s41586-021-03819-2
7. Economist T. Remarkable progress has been made in understanding the folding of proteins. *The Economist*. July 30 2021 2021;
8. The Nobel Prize. The Nobel Prize in Chemistry. 2024. Accessed 9 October, 2024. https://www.nobelprize.org/prizes/chemistry/
9. Schneider J, Meske C, Kuss P. Foundation Models A New Paradigm for Artificial Intelligence. *Business & Information Systems Engineering*. 2024/04/01 2024;66(2):221-231. doi:10.1007/s12599-024-00851-0
10. Vaswani A, Shazeer N, Parmar N, et al. Attention is all you need. *Advances in neural information processing systems*. 2017;30
11. Brown TB. Language models are few-shot learners. *arXiv preprint arXiv:200514165*. 2020;
12. Thirunavukarasu AJ, Ting DSJ, Elangovan K, Gutierrez L, Tan TF, Ting DSW. Large language models in medicine. *Nat Med*. Aug 2023;29(8):1930-1940. doi:10.1038/s41591-023-02448-8
13. OpenAI. Introducing ChatGPT. https://openai.com/blog/chatgpt
14. Zhao WX, Zhou K, Li J, et al. A survey of large language models. *arXiv preprint arXiv:230318223*. 2023;
15. National Institute for Health and Care Excellence. Use of AI in evidence generation: NICE position statement. 2024. Accessed 20 September, 2024. https://www.nice.org.uk/about/what-we-do/our-research-work/use-of-ai-in-evidence-generation--nice-position-statement
16. National Institute for Health and Care Excellence. NICE statement of intent for artificial intelligence (AI) 16 December, 2024. Accessed 16 December 2024. https://www.nice.org.uk/corporate/ecd12/resources/nice-statement-of-intent-for-artificial-intelligence-ai-pdf-40464270623941

17.	Tsertsvadze A, Chen YF, Moher D, Sutcliffe P, McCarthy N. How to conduct systematic reviews more expeditiously? *Syst Rev*. Nov 12 2015;4:160. doi:10.1186/s13643-015-0147-7

18.	*Cochrane Handbook for Systematic Reviews of Interventions. 2nd Edition. Chichester (UK): John Wiley & Sons, 2019.* Higgins JPT, Thomas J, Chandler J, Cumpston M, Li T, Page MJ, Welch VA (editors). John Wiley & Sons; 2019.

19.	Qureshi R, Shaughnessy D, Gill KAR, Robinson KA, Li T, Agai E. Are ChatGPT and large language models "the answer" to bringing us closer to systematic review automation? *Syst Rev*. Apr 29 2023;12(1):72. doi:10.1186/s13643-023-02243-z

20.	Khraisha Q, Put S, Kappenberg J, Warraitch A, Hadfield K. Can large language models replace humans in systematic reviews? Evaluating GPT-4's efficacy in screening and extracting data from peer-reviewed and grey literature in multiple languages. *Res Synth Methods*. Mar 14 2024;doi:10.1002/jrsm.1715

21.	Guo E, Gupta M, Deng J, Park YJ, Paget M, Naugler C. Automated Paper Screening for Clinical Reviews Using Large Language Models: Data Analysis Study. *J Med Internet Res*. Jan 12 2024;26:e48996. doi:10.2196/48996

22.	Tran VT, Gartlehner G, Yaacoub S, et al. Sensitivity and Specificity of Using GPT-3.5 Turbo Models for Title and Abstract Screening in Systematic Reviews and Meta-analyses. *Ann Intern Med*. Jun 2024;177(6):791-799. doi:10.7326/m23-3389

23.	Landschaft A, Antweiler D, Mackay S, et al. Implementation and evaluation of an additional GPT-4-based reviewer in PRISMA-based medical systematic literature reviews. *Int J Med Inform*. Sep 2024;189:105531. doi:10.1016/j.ijmedinf.2024.105531

24.	Robinson A, Thorne W, Wu BP, et al. Bio-sieve: Exploring instruction tuning large language models for systematic review automation. *arXiv preprint arXiv:230806610*. 2023;

25.	Hasan B, Saadi S, Rajjoub NS, et al. Integrating large language models in systematic reviews: a framework and case study using ROBINS-I for risk of bias assessment. *BMJ Evid Based Med*. Feb 21 2024;doi:10.1136/bmjebm-2023-112597

26.	Lai H, Ge L, Sun M, et al. Assessing the Risk of Bias in Randomized Clinical Trials With Large Language Models. *JAMA Netw Open*. May 1 2024;7(5):e2412687. doi:10.1001/jamanetworkopen.2024.12687

27.	Lee K, Paek H, Huang LC, et al. SEETrials: Leveraging Large Language Models for Safety and Efficacy Extraction in Oncology Clinical Trials. *medRxiv*. May 13 2024;doi:10.1101/2024.01.18.24301502

28.	Schopow N, Osterhoff G, Baur D. Applications of the Natural Language Processing Tool ChatGPT in Clinical Practice: Comparative Study and Augmented Systematic Review. *JMIR Med Inform*. Nov 28 2023;11:e48933. doi:10.2196/48933

29.	Gartlehner G, Kahwati L, Hilscher R, et al. Data extraction for evidence synthesis using a large language model: A proof-of-concept study. *Res Synth Methods*. Mar 3 2024;doi:10.1002/jrsm.1710

30.	Reason T, Benbow E, Langham J, Gimblett A, Klijn SL, Malcolm B. Artificial Intelligence to Automate Network Meta-Analyses: Four Case Studies to Evaluate the Potential

Application of Large Language Models. *Pharmacoecon Open*. Mar 2024;8(2):205-220. doi:10.1007/s41669-024-00476-9

31. Jansen JP, Fleurence R, Devine B, et al. Interpreting indirect treatment comparisons and network meta-analysis for health-care decision making: report of the ISPOR Task Force on Indirect Treatment Comparisons Good Research Practices: part 1. *Value Health*. Jun 2011;14(4):417-28. doi:10.1016/j.jval.2011.04.002

32. Yun HS, Pogrebitskiy D, Marshall IJ, Wallace BC. Automatically Extracting Numerical Results from Randomized Controlled Trials with Large Language Models. *arXiv preprint arXiv:240501686*. 2024;

33. Huang Y, Wu R, He J, Xiang Y. Evaluating ChatGPT-4.0's data analytic proficiency in epidemiological studies: A comparative analysis with SAS, SPSS, and R. *J Glob Health*. Mar 29 2024;14:04070. doi:10.7189/jogh.14.04070

34. Wu Y, Klijn S, Teitsson S, Malcolm B, Jones C, Rawlinson W. Innovations in Automated Survival Curve Selection and Reporting of Survival Analyses Through Generative AI. Presented at ISPOR Europe, Barcelona. . 2024, Accessed 19 January, 2025. https://www.ispor.org/heor-resources/presentations-database/presentation-paper/euro2024-4003/19093/innovations-in-automated-survival-curve-selection-and-reporting-of-survival-analyses-through-generative-ai

35. Jin Q, Leaman R, Lu Z. Retrieve, Summarize, and Verify: How Will ChatGPT Affect Information Seeking from the Medical Literature? *J Am Soc Nephrol*. Aug 1 2023;34(8):1302-1304. doi:10.1681/ASN.0000000000000166

36. Smela B, Łukiewicz B, Gawlik K, Clay E, Boyer L TM. Balancing Feasibility, Time, and Comprehensiveness: Approaches to Rapid Reviews of Health Economic Models. presented at: ISPOR 2024; June 2024 2024; Atlanta, GA, USA. https://www.ispor.org/heor-resources/presentations-database/presentation/intl2024-3896/138795

37. Chhatwal J, Yildrim IF, Balta D, et al. Can Large Language Models Generate Conceptual Health Economic Models? . presented at: ISPOR 2024; 2024; Atlanta, Georgia. https://www.ispor.org/heor-resources/presentations-database/presentation/intl2024-3898/139128

38. Chhatwal J, Yildirim IF, Samur S, Bayraktar E, Ermis T, T A. Development of De Novo Health Economic Models Using Generative AI. presented at: ISPOR Europe 2024 Meeting; 2024; Barcelona, Spain. https://www.valueinhealthjournal.com/article/S1098-3015(24)02899-7/abstract

39. Reason T, Rawlinson W, Langham J, Gimblett A, Malcolm B, Klijn S. Artificial Intelligence to Automate Health Economic Modelling: A Case Study to Evaluate the Potential Application of Large Language Models. *Pharmacoecon Open*. Mar 2024;8(2):191-203. doi:10.1007/s41669-024-00477-8

40. Chhatwal J, Samur S, Yildirim IF, Bayraktar E, Ermis T, Ayer T. Fully Replicating Published Health Economic Models Using Generative AI. presented at: ISPOR Europe 2024 Meeting; 2024; Barcelona, Spain. https://www.valueinhealthjournal.com/article/S1098-3015(24)03392-8/abstract

41. Briggs AH, Weinstein MC, Fenwick EA, Karnon J, Sculpher MJ, Paltiel AD. Model parameter estimation and uncertainty: a report of the ISPOR-SMDM Modeling Good Research Practices Task Force--6. *Value Health*. Sep-Oct 2012;15(6):835-42. doi:10.1016/j.jval.2012.04.014
42. Bedi S, Liu Y, Orr-Ewing L, et al. Testing and Evaluation of Health Care Applications of Large Language Models: A Systematic Review. *JAMA*. 2024;doi:10.1001/jama.2024.21700
43. Fleurence RL, Dawoud D, Bian J, et al. The ELEVATE-AI LLMs Framework: An Evaluation Framework for Use of Large Language Models in HEOR: an ISPOR Working Group Report. *arXiv:250112394*. 2024;
44. Fleurence R, Kent S, Adamson B, et al. Assessing Real-World Data from Electronic Health Records for Health Technology Assessment – The SUITABILITY Checklist: A Good Practices Report from an ISPOR Task Force. *Value in Health*. 2024;27(6):692-701.
45. Lee K, Liu Z, Chandran U, et al. Detecting Ground Glass Opacity Features in Patients With Lung Cancer: Automated Extraction and Longitudinal Analysis via Deep Learning–Based Natural Language Processing. *JMIR AI*. 2023/6/1 2023;2:e44537. doi:10.2196/44537
46. Hu Y, Chen Q, Du J, et al. Improving large language models for clinical named entity recognition via prompt engineering. *Journal of the American Medical Informatics Association*. 2024;31(9):1812-1820. doi:10.1093/jamia/ocad259
47. Guo LL, Fries J, Steinberg E, et al. A multi-center study on the adaptability of a shared foundation model for electronic health records. *npj Digital Medicine*. 2024/06/27 2024;7(1):171. doi:10.1038/s41746-024-01166-w
48. Cohen AB, Waskom M, Adamson B, Kelly J, G A. Using Large Language Models To Extract PD-L1 Testing Details From Electronic Health Records. presented at: ISPOR 2024; 2024; Atlanta, GA. https://www.ispor.org/heor-resources/presentations-database/presentation/intl2024-3898/136019
49. Soroush A, Glicksberg BS, Zimlichman E, et al. Large Language Models Are Poor Medical Coders — Benchmarking of Medical Code Querying. *NEJM AI*. 2024;1(5):AIdbp2300040. doi:doi:10.1056/AIdbp2300040
50. Peng C, Yang X, Chen A, et al. A study of generative large language model for medical research and healthcare. *npj Digital Medicine*. 2023/11/16 2023;6(1):210. doi:10.1038/s41746-023-00958-w
51. Jiang LY, Liu XC, Nejatian NP, et al. Health system-scale language models are all-purpose prediction engines. *Nature*. 2023/07/01 2023;619(7969):357-362. doi:10.1038/s41586-023-06160-y
52. Yang X, Chen A, PourNejatian N, et al. A large language model for electronic health records. *NPJ Digit Med*. Dec 26 2022;5(1):194. doi:10.1038/s41746-022-00742-2
53. Xie Q, Chen Q, Chen A, et al. Me-LLaMA: Foundation Large Language Models for Medical Applications. *Res Sq*. May 22 2024;doi:10.21203/rs.3.rs-4240043/v1
54. Yang L, Xu S, Sellergren A, et al. Advancing Multimodal Medical Capabilities of Gemini. *arXiv preprint arXiv:240503162*. 2024;

55. Rajpurkar P, Chen E, Banerjee O, Topol EJ. AI in health and medicine. *Nat Med*. Jan 2022;28(1):31-38. doi:10.1038/s41591-021-01614-0
56. Du H, Zhao J, Zhao Y, et al. Advancing Real-time Pandemic Forecasting Using Large Language Models: A COVID-19 Case Study. *arXiv preprint arXiv:240406962*. 2024;
57. Smith GR, Bello C, Bialic-Murphy L, et al. Ten simple rules for using large language models in science, version 1.0. *PLoS Comput Biol*. Jan 2024;20(1):e1011767. doi:10.1371/journal.pcbi.1011767
58. Team Gemini, Anil R, Borgeaud S, et al. Gemini: a family of highly capable multimodal models. *arXiv preprint arXiv:231211805*. 2023;
59. August T, Lo K, Smith NA, Reinecke K. Know Your Audience: The benefits and pitfalls of generating plain language summaries beyond the" general" audience. 2024:1-26.
60. Sivarajkumar S, Kelley M, Samolyk-Mazzanti A, Visweswaran S, Wang Y. An Empirical Evaluation of Prompting Strategies for Large Language Models in Zero-Shot Clinical Natural Language Processing: Algorithm Development and Validation Study. *JMIR Med Inform*. 2024/4/8 2024;12:e55318. doi:10.2196/55318
61. Schulhoff S, Ilie M, Balepur N, et al. The Prompt Report: A Systematic Survey of Prompting Techniques. *arXiv preprint arXiv:240606608*. 2024;
62. Lin Z. How to write effective prompts for large language models. *Nat Hum Behav*. Mar 4 2024;doi:10.1038/s41562-024-01847-2
63. Lin Z. Why and how to embrace AI such as ChatGPT in your academic life. *R Soc Open Sci*. Aug 2023;10(8):230658. doi:10.1098/rsos.230658
64. Kojima T, Shane Gu S, Reid M, Matsuo Y, Iwasawa Y. Large language models are zero-shot reasoners. presented at: Proceedings of the 36th International Conference on Neural Information Processing Systems; 2024; New Orleans, LA, USA. https://dl.acm.org/doi/10.5555/3600270.3601883
65. Wei J, Wang X, Schuurmans D, et al. Chain-of-thought prompting elicits reasoning in large language models. *Advances in neural information processing systems*. 2022;35:24824-24837.
66. Chung HW, Hou L, Longpre S, et al. Scaling instruction-finetuned language models. *Journal of Machine Learning Research*. 2024;25(70):1-53.
67. Ouyang L. Training language models to follow instructions with human feedback. 2022;
68. Yue X, Zheng T, Zhang G, Chen W. Mammoth2: Scaling instructions from the web. *arXiv preprint arXiv:240503548*. 2024;
69. Casper S, Davies X, Shi C, et al. Open problems and fundamental limitations of reinforcement learning from human feedback. *arXiv preprint arXiv:230715217*. 2023;
70. Szép M, Rueckert D, von Eisenhart-Rothe R, Hinterwimmer F. A Practical Guide to Fine-tuning Language Models with Limited Data. *arXiv preprint arXiv:241109539*. 2024;
71. Moor M, Banerjee O, Abad ZSH, et al. Foundation models for generalist medical artificial intelligence. *Nature*. Apr 2023;616(7956):259-265. doi:10.1038/s41586-023-05881-4

72. Alsentzer E, Murphy JR, Boag W, et al. Publicly available clinical BERT embeddings. *arXiv preprint arXiv:190403323*. 2019;
73. Gao Y, Xiong Y, Gao X, et al. Retrieval-augmented generation for large language models: A survey. *arXiv preprint arXiv:231210997*. 2023;
74. Lewis P, Perez E, Piktus A, et al. Retrieval-augmented generation for knowledge-intensive NLP tasks. presented at: Proceedings of the 34th International Conference on Neural Information Processing Systems; 2020; Vancouver, BC, Canada.
75. OpenAI. Retrieval Augmented Generation (RAG) and Semantic Search for GPTs. Accessed 20 January 2025. https://help.openai.com/en/articles/8868588-retrieval-augmented-generation-rag-and-semantic-search-for-gpts
76. Lin XV, Chen X, Chen M, et al. Ra-dit: Retrieval-augmented dual instruction tuning. *arXiv preprint arXiv:231001352*. 2023;
77. Yu H, Gan A, Zhang K, Tong S, Liu Q, Liu Z. Evaluation of Retrieval-Augmented Generation: A Survey. *arXiv preprint arXiv:240507437*. 2024;
78. Xi Z, Chen W, Guo X, et al. The rise and potential of large language model based agents: A survey. *arXiv preprint arXiv:230907864*. 2023;
79. Cheng Y, Zhang C, Zhang Z, et al. Exploring large language model based intelligent agents: Definitions, methods, and prospects. *arXiv preprint arXiv:240103428*. 2024;
80. Clarke C, Krishnamurthy K, Talamonti W, Kang Y, Tang L, Mars J. One Agent Too Many: User Perspectives on Approaches to Multi-agent Conversational AI. *arXiv preprint arXiv:240107123*. 2024;
81. Langchain. Accessed 20 January, 2025. https://www.langchain.com/
82. Microsoft. AutogenAutoGen: Open-Source Programming Framework for Agentic AI. Accessed 20 January, 2025. https://www.microsoft.com/en-us/research/project/autogen/
83. GPT Researcher. GPT Researcher. Accessed 20 January 2025. https://gptr.dev/
84. Beam AL, Manrai AK, Ghassemi M. Challenges to the Reproducibility of Machine Learning Models in Health Care. *JAMA*. Jan 28 2020;323(4):305-306. doi:10.1001/jama.2019.20866
85. Kapoor S, Cantrell EM, Peng K, et al. REFORMS: Consensus-based Recommendations for Machine-learning-based Science. *Sci Adv*. May 3 2024;10(18):eadk3452. doi:10.1126/sciadv.adk3452
86. Wei CH, Allot A, Lai PT, et al. PubTator 3.0: an AI-powered literature resource for unlocking biomedical knowledge. *Nucleic Acids Res*. Apr 4 2024;doi:10.1093/nar/gkae235
87. Xie Q, Chen Q, Chen A, et al. Me LLaMA: Foundation Large Language Models for Medical Applications. *arXiv preprint arXiv:240212749*. 2024;
88. Caton S, Haas C. Fairness in machine learning: A survey. *ACM Computing Surveys*. 2024;56(7):1-38.
89. Mehrabi N, Morstatter F, Saxena N, Lerman K, Galstyan A. A survey on bias and fairness in machine learning. *ACM computing surveys (CSUR)*. 2021;54(6):1-35.

90. Drukker K, Chen W, Gichoya J, et al. Toward fairness in artificial intelligence for medical image analysis: identification and mitigation of potential biases in the roadmap from data collection to model deployment. *J Med Imaging (Bellingham)*. Nov 2023;10(6):061104. doi:10.1117/1.JMI.10.6.061104

91. Yang Y, Lin M, Zhao H, Peng Y, Huang F, Lu Z. A survey of recent methods for addressing AI fairness and bias in biomedicine. *J Biomed Inform*. Apr 25 2024:104646. doi:10.1016/j.jbi.2024.104646

92. Gervasi S, Chen I, Smith-McLallen A, et al. The Potential For Bias In Machine Learning And Opportunities For Health Insurers To Address It. *Health Affairs*. 2022;41(2):212-218. doi:10.1377/hlthaff.2021.01287

93. Gichoya JW, Banerjee I, Bhimireddy AR, et al. AI recognition of patient race in medical imaging: a modelling study. *Lancet Digit Health*. Jun 2022;4(6):e406-e414. doi:10.1016/S2589-7500(22)00063-2

94. Xu J, Xiao Y, Wang WH, et al. Algorithmic fairness in computational medicine. *EBioMedicine*. Oct 2022;84:104250. doi:10.1016/j.ebiom.2022.104250

95. Huang Y, Guo J, Chen WH, et al. A scoping review of fair machine learning techniques when using real-world data. *J Biomed Inform*. Mar 2024;151:104622. doi:10.1016/j.jbi.2024.104622

96. Mosquera L, El Emam K, Ding L, et al. A method for generating synthetic longitudinal health data. *BMC Med Res Methodol*. Mar 23 2023;23(1):67. doi:10.1186/s12874-023-01869-w

97. European Union. EU Artificial Intelligence Act Accessed May 22, 2024. https://artificialintelligenceact.eu/the-act/

98. Benitez K, Malin B. Evaluating re-identification risks with respect to the HIPAA privacy rule. *J Am Med Inform Assoc*. Mar-Apr 2010;17(2):169-77. doi:10.1136/jamia.2009.000026

99. Simon GE, Shortreed SM, Coley RY, et al. Assessing and Minimizing Re-identification Risk in Research Data Derived from Health Care Records. *EGEMS (Wash DC)*. Mar 29 2019;7(1):6. doi:10.5334/egems.270

100. HITRUST. Accessed 20 January, 2025. https://hitrustalliance.net/hitrust-framework

101. FISMA. Accessed 20 January, 2025. https://csrc.nist.gov/topics/laws-and-regulations/laws/FISMA

102. Ren C, Yu H, Peng H, et al. Advances and open challenges in federated learning with foundation models. *arXiv e-prints*. 2024:arXiv: 2404.15381.

103. Lu S. Proprietary vs. Open Source Foundation Models. Accessed 20 January, 2025. https://tolacapital.com/2023/05/15/foundationmodels?utm_source=chatgpt.com

104. Luqman A, Mahesh R, Chattopadhyay A. Privacy and security implications of cloud-based ai services: A survey. *arXiv preprint arXiv:240200896*. 2024;

105. Zandesh Z. Privacy, Security, and Legal Issues in the Health Cloud: Structured Review for Taxonomy Development. *JMIR Form Res*. Feb 12 2024;8:e38372. doi:10.2196/38372

106. Zemplenyi A, Tachkov K, Balkanyi L, et al. Recommendations to overcome barriers to the use of artificial intelligence-driven evidence in health technology assessment. *Front Public Health*. 2023;11:1088121. doi:10.3389/fpubh.2023.1088121

107. Jordan MI, Mitchell TM. Machine learning: Trends, perspectives, and prospects. *Science*. Jul 17 2015;349(6245):255-60. doi:10.1126/science.aaa8415

108. Breiman L. Random Forests. *Machine Learning*. 2001/10/01 2001;45(1):5-32. doi:10.1023/A:1010933404324

109. Vapnik V. *The Nature of Statistical Learning Theory*. Springer-Verlag; 1995.

110. Padula WV, Kreif N, Vanness DJ, et al. Machine Learning Methods in Health Economics and Outcomes Research-The PALISADE Checklist: A Good Practices Report of an ISPOR Task Force. *Value Health*. Jul 2022;25(7):1063-1080. doi:10.1016/j.jval.2022.03.022
27

**Table 1: Applications of generative AI to areas of HEOR**

| Area of HEOR | Potential and Actual Areas of Use |
|---|---|
| Systematic Literature Reviews | Search strategy, abstract screening, full text screening, bias assessment, data extraction, meta-analyses, report writing. |
| Economic Modeling | Economic model literature summarization, model conceptualization, parameter generation, code generation, structural uncertainty analysis, report writing. |
| Real-World Evidence | Data extraction, information retrieval, transformation of unstructured data to structured data, integration of multimodal data. |
| Dossier Development | Report writing in different styles and formats. |

**Table 2: Approaches to improve the quality of of generative AI outputs in HEOR**

| Technique/Approach | Description | Examples/Applications |
|---|---|---|
| Prompt Engineering | Methods to refine input prompts for generative AI models to enhance output accuracy and comprehensiveness | Examples include zero-shot, one-shot learning and chain-of-thought prompting to solve multi-step problems or more complex scenarios. |
| Retrieval-Augmented Generation (RAG) | Models augment their generative process by retrieving external, domain-specific knowledge to improve accuracy and factuality. | Integration of any data sources (e.g., publications, reports, images) for answering specific queries; Bing or ChatGPT querying external websites for fact-checking and validation. |
| Model Fine-Tuning | Adjusting pre-trained models on specific datasets for enhanced performance in specialized tasks to improve accuracy and completeness. | Fine-tuning on proprietary datasets for custom task performance (e.g., domain-specific text like clinical EHRs); use of reinforcement learning with human feedback (RLHF) to improve factual accuracy. |
| Domain-Specific FMs | Development and application of models trained on domain-specific corpora for increased accuracy and comprehensiveness. | Examples include GatorTron or BioClinicalBERT for specialized medical language understanding. |
| Multiple Agents | Deploying distinct agents specialized in different tasks to work collaboratively. This will increase productivity and speed. | An example would be a retrieval agent for sourcing data or references, a summarization agent to condense information, and an analysis agent for data-driven insights, each working in tandem to solve complex problems. |